\documentclass{article}

\PassOptionsToPackage{numbers, compress}{natbib}

\usepackage[preprint]{neurips_2022}

\usepackage[utf8]{inputenc} 
\usepackage[T1]{fontenc}    
\usepackage{hyperref}       
\usepackage{url}            
\usepackage{booktabs}       
\usepackage{amsfonts}       
\usepackage{nicefrac}       
\usepackage{microtype}      
\usepackage{xcolor}         
\usepackage{amsmath, amssymb}
\usepackage{subfigure}
\usepackage[pdftex]{graphicx}
\usepackage{multirow}

\usepackage{bm}

\newcommand{\CM}{\mathcal{M}}

\newcommand{\BF}{\mathbf{F}}
\newcommand{\bx}{\bm{x}}

\newcommand{\CF}{\mathcal{F}}

\newcommand{\CV}{\mathcal{V}}
\newcommand{\CT}{\mathcal{T}}
\newcommand{\CA}{\mathcal{A}}
\newcommand{\ba}{\bm{a}}
\newcommand{\bb}{\bm{b}}
\newcommand{\bc}{\bm{c}}

\title{Exploring Spatial-Temporal Features for \\ Deepfake Detection and Localization}

\author{
	Haiwei Wu \\
	University of Macau\\
	\And
	Jiantao Zhou* \\
	University of Macau\\
	\And
	Shile Zhang \\
	University of Macau\\
	\And
	Jinyu Tian \\
	Macau University of Science and Technology\\
}

\begin{document}
	
\maketitle

\begin{abstract}
	With the continuous research on Deepfake forensics, recent studies have attempted to provide the fine-grained localization of forgeries, in addition to the coarse classification at the video-level. However, the detection and localization performance of existing Deepfake forensic methods still have plenty of room for further improvement. In this work, we propose a Spatial-Temporal Deepfake Detection and Localization (ST-DDL) network that simultaneously explores spatial and temporal features for detecting and localizing forged regions. Specifically, we design a new Anchor-Mesh Motion (AMM) algorithm to extract temporal (motion) features by modeling the precise geometric movements of the facial micro-expression. Compared with traditional motion extraction methods (e.g., optical flow) designed to simulate large-moving objects, our proposed AMM could better capture the small-displacement facial features. The temporal features and the spatial features are then fused in a Fusion Attention (FA) module based on a Transformer architecture for the eventual Deepfake forensic tasks. The superiority of our ST-DDL network is verified by experimental comparisons with several state-of-the-art competitors, in terms of both video- and pixel-level detection and localization performance. Furthermore, to impel the future development of Deepfake forensics, we build a public forgery dataset consisting of 6000 videos, with many new features such as using widely-used commercial software (e.g., After Effects) for the production, providing online social networks transmitted versions, and splicing multi-source videos. The source code and dataset are available at \textbf{https://github.com/HighwayWu/ST-DDL}.
\end{abstract}

\section{Introduction}
The research of deep generative models has greatly promoted the development of human face synthetic techniques, e.g., identity swap and attributes manipulation. Although the synthesized media (i.e., Deepfake) could contribute to the film and entertainment industry \cite{face2face, fsgan, aot}, they are becoming increasingly dangerous in various fields such as stealing identity, modifying facial properties, and producing celebrity pornography, negatively affecting not only individuals but also the whole society. Therefore, it is extremely urgent to develop forensic algorithms to detect Deepfake videos, preferably with capabilities of localizing forged regions at pixel-level accuracy.

Many of the existing methods \cite{xray, fourier2020, what2020, local2021, facecontext, selfconsistency} (and references therein) were committed to detecting whether a Deepfake video is pristine or synthesized. Some works tried to mine forgery traces in the spatial domain, such as blending boundaries \cite{xray}, discrepancies between faces and their context \cite{facecontext}, and inconsistencies of source features within the manipulated faces \cite{selfconsistency}. Aiming at revealing fine-grained forensic clues, some algorithms \cite{nguyen2019multi, fakelocator, wavenet, cffs} have been recently proposed to localize the forged regions in Deepfake videos, rather than simply offering binary classification results (pristine or synthesized). Specifically, FakeLocator \cite{fakelocator} first attempted to solve the fake localization problem by depicting the imperfection of the upsampling operation in the GAN-based forgeries. To further improve the generalization performance to unseen manipulations, CFFs \cite{cffs} proposed a commonality learning strategy to learn common forgery traces from different databases. Note that these algorithms \cite{xray, fourier2020, what2020, local2021, facecontext, selfconsistency, nguyen2019multi, fakelocator, wavenet, cffs} solely exploited the spatial information, while neglecting the valuable temporal information, which could significantly boost the forgery detection and localization performance \cite{gu2021spatiotemporal, ftcn2021}.

\begin{figure}[t]
	\centering
	\subfigure{
		\includegraphics[width = 0.98\textwidth]{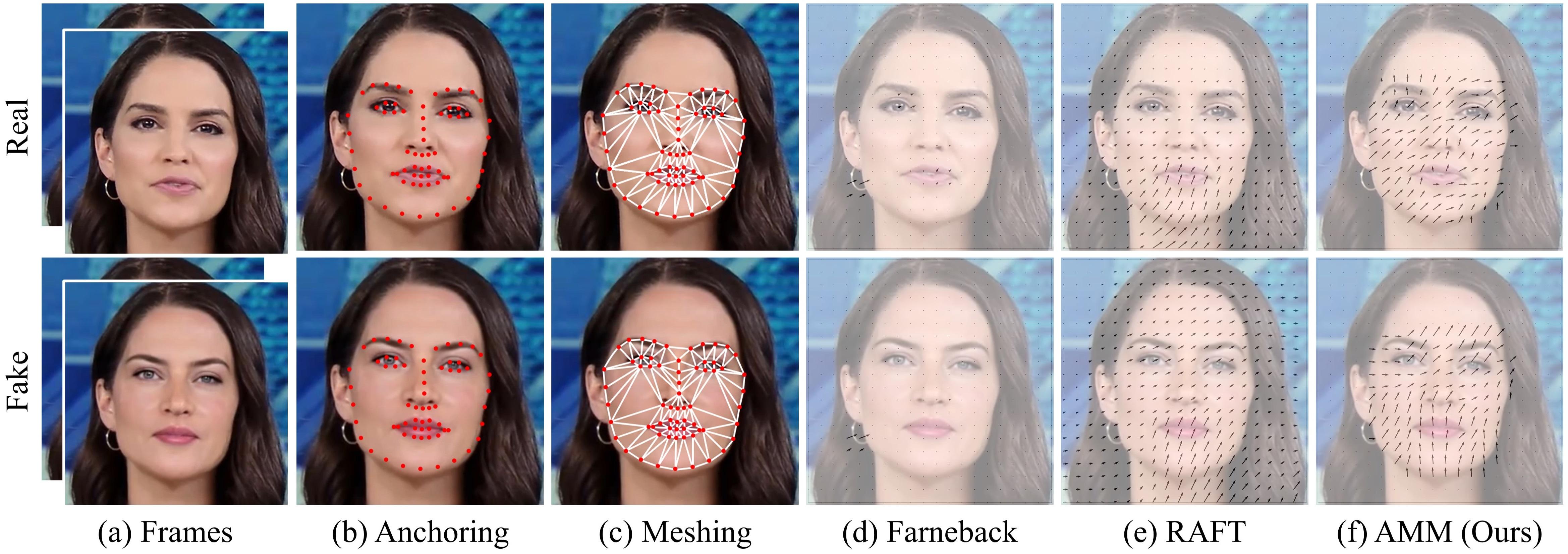}}
	\caption{Motions extracted from real and fake faces by (d) Farneback \cite{farneback}, (e) RAFT \cite{raft} and (f) our AMM, respectively. As can be observed, the authenticity is difficult to distinguish from the Farneback and RAFT motions, where only minimal or even homogeneous temporal features are provided. In contrast, by anchoring and gridding the faces, our AMM characterizes the facial movements more delicately, providing guidance for the subsequent forgery detection and localization. Noted that the (d)-(e) motions are directly extracted from (a), while ours are from (c).} 
	\label{fig:motion_cmp}
\end{figure}

The temporal information plays an important role in many fields (e.g., video prediction \cite{VideoPrediction}, video super-resolution \cite{videosr}, and pose estimation \cite{PoseEstimation}), and hence, it is natural to seek for additional clues in the temporal domain to enhance the Deepfake forensics. Along this line, some recent attempts \cite{flowcnn, lip2021, ftcn2021, chugh2020not, chen2020fsspotter, zhang2021detecting, gu2021spatiotemporal, cozzolino2021id, li2021deepfake, sun2021improving} explored temporal inconsistencies from fake videos and incorporated with spatial evidence. The pioneering work \cite{flowcnn} utilized the global motion feature \cite{pwcnet} to reveal the dissimilarities between consecutive frames. Later, LipNet \cite{lip2021} focused on analyzing high-level semantic irregularities in mouth movements. By analyzing the motion inconsistency along horizontal and vertical directions, STIL \cite{gu2021spatiotemporal} depicted spatial-temporal discontinuous burrs for Deepfake detection. Here, we would like to emphasize that, among all Deepfake forensic methods leveraging spatial-temporal features, the most crucial task is to design these features appropriately, and hence, distinct designs could lead to dramatically different performance. For example, we find that the existing motion algorithms \cite{farneback, pwcnet, raft} may be uncomfortably applied to Deepfake forensics. As shown in Fig.~\ref{fig:motion_cmp} (d), the traditional algorithm Farneback \cite{farneback} sometimes fails to extract tiny movements, while the motion extracted by the latest deep learning method RAFT \cite{raft} is homogeneous (see Fig.~\ref{fig:motion_cmp} (e)). In other words, the motions extracted by existing methods \cite{farneback, pwcnet, raft} cannot model the delicate changes of facial expressions, which are important for detecting and localizing Deepfake videos.

In this work, we propose a novel Spatial-Temporal Deepfake Detection and Localization (ST-DDL) network that simultaneously explores spatial and temporal features for detecting and localizing forged regions in Deepfake videos. To extract temporal features dedicated to Deepfake forensic tasks, we design an Anchor-Mesh Motion (AMM) algorithm. More specifically, AMM formulates the homeomorphism mapping between the faces in adjacent frames. However, if such a mapping is optimized over the entire facial domain, it is not only computationally expensive to seek the corresponding points in contiguous frames, but also inefficient to generate dense motions for the entire face. Inspired by the observation that facial landmarks usually indicate the ideal face location in the Deepfake generation, we naturally serve the landmark displacements as pre-defined motions. Then we propose to triangularly grid the face by utilizing the landmarks as anchors (Fig. \ref{fig:motion_cmp} (b)-(c)), so as to approximate the mapping optimization by more efficient triangular transformation. An example of the AMM motion is shown in Fig.~\ref{fig:motion_cmp} (f), which clearly presents more discriminative clues between real and fake faces. In addition, to further promote the interactivity of spatial-temporal (RGB and motion) features in the learning process, we introduce a Fusion Attention (FA) module into the ST-DDL, which is built upon the powerful self-attention mechanism \cite{attention}. Experimental comparisons with the state-of-the-art approaches \cite{flowcnn, lip2021, ftcn2021, wavenet, fakelocator, cffs} validate the effectiveness of our method, increasing the detection accuracy (video-level F1) by up to 8.9\% and the localization accuracy (pixel-level IoU) by 4.1\%.

Noteworthy, datasets produced by state-of-the-art forgery methods could greatly advance forensic research \cite{aot}. However, existing datasets \cite{UADFV, TIMIT, dfd, DFFD, ff++, DFDC, CelebDF, DeeperForensics, ForgeryNet, ffiw} either have rather limited diversity of forgery categories or lack more realistic test scenarios. To remedy these deficiencies, we construct a new Deepfake dataset, called \texttt{ManualFake}, consisting of 6000 videos, with many new features such as using widely-used commercial software (e.g., After Effects) for the production, providing online social networks (OSNs) transmitted versions, and splicing multi-source videos (e.g., online interviews).


In summary, our major contributions are: (1) By anchoring and gridding the face, we propose an AMM algorithm to mine more distinctive spatial-temporal features for Deepfake forensics. (2) We further propose a FA module to facilitate the attention learning of latent spatial-temporal forensic features, realizing a ST-DDL network. Extensive comparisons with several state-of-the-art methods \cite{flowcnn, lip2021, ftcn2021, wavenet, fakelocator, cffs} demonstrate the superiority of our network, in terms of both video-level detection and pixel-level localization performance. (3) We build a public forgery dataset, \texttt{ManualFake}, for future research in the Deepfake forensics community. To the best of our knowledge, this is the \textit{first} Deepfake dataset with unique properties, e.g., using commercial software for the production, providing OSN-transmitted versions, and splicing multi-source videos.


	

\section{Spatial-Temporal Deepfake Detection and Localization}
\begin{figure}[t]
	\centering
	\subfigure{
		\includegraphics[width = 0.9\textwidth]{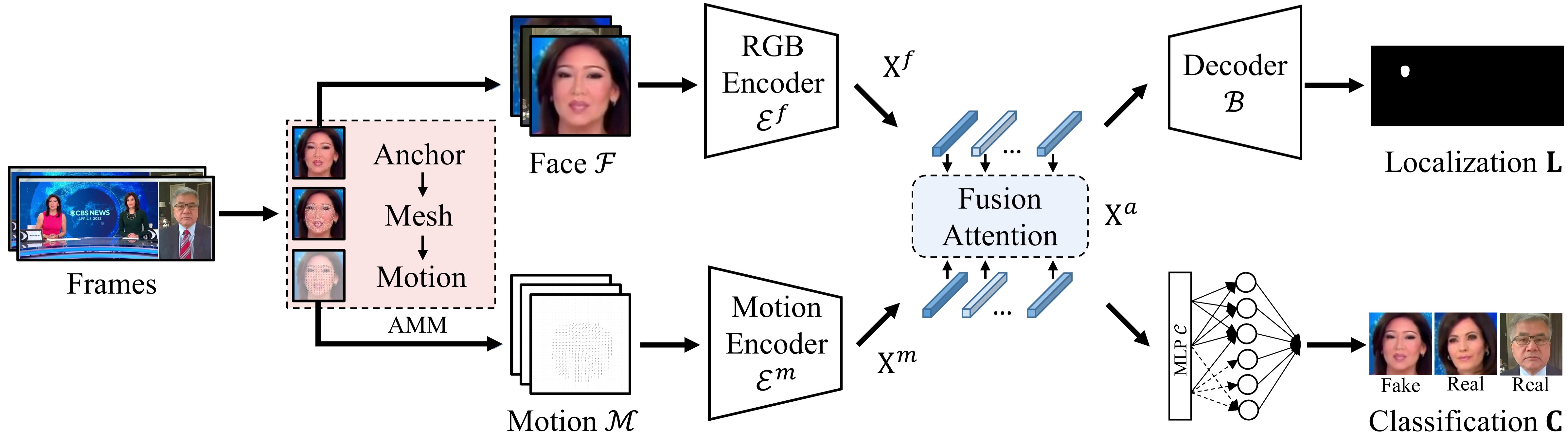}}
	\caption{The overall framework of the Spatial-Temporal Deepfake Detection and Localization (ST-DDL) network. Given a video under investigation, ST-DDL produces both video-level detection and pixel-level localization results.}
	\label{fig:framework}
\end{figure}

The schematic diagram of our ST-DDL network is illustrated in Fig. \ref{fig:framework}. After cropping and aligning a series of faces from a given video under investigation, our model first performs the AMM algorithm to extract motions. Then the original faces and extracted motions are fed separately into two encoders; one for learning features from RGB (spatial) domain, and the other from motion (temporal) domain. As expected and will be verified experimentally, exploring spatial-temporal features simultaneously could significantly improve the Deepfake detection and localization performance. Upon obtaining the preliminarily encoded features, we utilize the FA module to enhance their informative interactions. Finally, a decoder and a multi-layer perceptron (MLP) are employed to output specific localization and classification results, respectively. In the following Sec. \ref{sec:fusion}, we first explain the overall network architecture as well as the FA module. We then give the details of the AMM algorithm in Sec. \ref{sec:motion}. To simplify the subsequent presentation, we assume that there is only one face in each video frame. Our method can also be readily extended to more general cases with multiple faces per frame.

\subsection{Network Architecture and Fusion Attention}\label{sec:fusion}
Many different encoders have been proposed for the feature extraction, such as ResNet \cite{resnet2016}, EfficientNet \cite{efficientnet} and High-Resolution Network (HRNet) \cite{hrnet}, each with unique application scenarios. In our ST-DDL network, we adopt the HRNet \cite{hrnet} as the RGB- and motion-domain encoders. Such a selection is based on a preliminary experiment to compare the performance of different encoders for the specific Deepfake detection and localization tasks. Another critical problem is how to fuse the extracted features from RGB and motion domains. Clearly, a naive convolutional concatenation strategy could be inferior to effectively explore these features. Inspired by the powerful Transformer architecture \cite{attention, chen2022activating}, we incorporate a trainable linear projection module, termed Fusion Attention (FA), to map the features from the motion branch to the RGB branch. 

\begin{figure}[t]
	\centering
	\subfigure{
		\includegraphics[width = 0.75\textwidth]{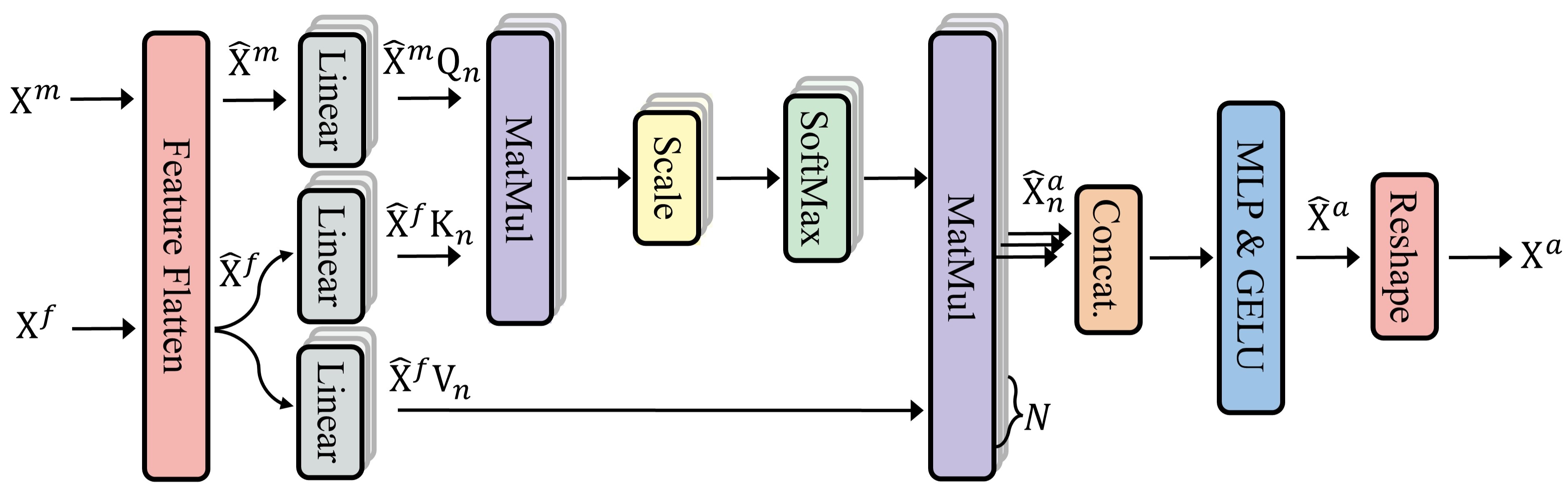}}
	\caption{The architecture of the FA module. }
	\label{fig:fuse}
	\vspace{-2mm}
\end{figure}

Specifically, for a given video with $I$ frames, we first pre-process it and generate a series of face images $\{\CF_i\}_{i=1}^I \in \mathbb{R}^{H \times W \times 3}$, by using an existing face recognition algorithm (e.g., RetinaFace \cite{retinaface}). Then the motion $\CM_i \in \mathbb{R}^{H \times W \times 2}$ is derived for each pair of adjacent faces $\CF_i$ and $\CF_{i+1}$ by our proposed AMM algorithm. The discussion on the AMM algorithm is deferred to the next subsection. The encoders $\mathcal{E}^f$ and $\mathcal{E}^m$ take $\CF_i$ and $\CM_i$ as inputs, and extract the deep latent features $\mathbf{X}^f$ and $\mathbf{X}^m$, respectively. Formally, we have 
\begin{equation}
    \mathbf{X}^f = \mathcal{E}^f(\CF_i),~~\mathbf{X}^m = \mathcal{E}^m(\CM_i).
\end{equation}
To learn the more discriminative features jointly from both RGB and motion domains, instead of simply using concatenation and convolution operations, we adopt the newly proposed FA mapping    
\begin{equation}\label{eq:fuse}
\mathbf{X}^a = \mathrm{FA}(\mathbf{X}^f, \mathbf{X}^m).
\end{equation}
The structure of FA is depicted in Fig. \ref{fig:fuse}. Specifically, given a latent feature $\mathbf{X}^f \in \mathbb{R}^{\hat{H} \times \hat{W} \times C}$, we first perform tokenization by flattening it into $\hat{\mathbf{X}}^f \in \mathbb{R}^{\hat{H}\hat{W} \times C}$. The similar operations are conducted on $\mathbf{X}^m$ to generate $\hat{\mathbf{X}}^m$. Then we perform $N$-head attention on the flattened features $\hat{\mathbf{X}}^f$ and $\hat{\mathbf{X}}^m$ by calculating attention on every $d = C/N$ channels, obtaining the internal features $\{\hat{\mathbf{X}}^a_n\}_{n=1}^N$, where
\begin{equation}
	\hat{\mathbf{X}}^a_n = \mathrm{Attention}(\hat{\mathbf{X}}^m\mathbf{Q}_n, \hat{\mathbf{X}}^f\mathbf{K}_n, \hat{\mathbf{X}}^f\mathbf{V}_n), n=1, ..., N,
\end{equation} and $\mathbf{Q}_n, \mathbf{K}_n, \mathbf{V}_n \in \mathbb{R}^{C \times d}$ are \textit{query}, \textit{key}, and \textit{value} matrices of the $n$-th projection for the self-attention function \cite{attention}. Here, the self-attention is performed by 
\begin{equation}
	\mathrm{Attention}(\mathbf{Q}, \mathbf{K}, \mathbf{V}) = \mathrm{SoftMax}(\frac{\mathbf{Q}\mathbf{K}^T}{\sqrt{d}})\textbf{V}.
\end{equation}
Next, the concatenated outputs of all heads $\{\hat{\mathbf{X}}^a_n\}_{n=1}^N$ are linearly projected to generate the flattened attention $\hat{\mathbf{X}}^a$ via
\begin{equation}
    \hat{\mathbf{X}}^a = \mathrm{MLP}(\mathrm{Concat}(\mathbf{X}^a_1, \mathbf{X}^a_2, ..., \mathbf{X}^a_N)),
\end{equation}
in which $\mathrm{MLP}(\cdot)$ represents a MLP with GELU activation \cite{gelu}. Finally, the fusion attention $\mathbf{X}^a$ is obtained by reshaping $\hat{\mathbf{X}}^a$ into $\hat{H} \times \hat{W} \times C$ resolutions. 

Regarding the decoder structure in Fig. \ref{fig:framework}, we follow the convention \cite{xray} and combine a series of blocks to form $\mathcal{B}: \mathbb{R}^{\hat{H} \times \hat{W} \times C} \to \mathbb{R}^{H \times W \times 1}$, where each block is composed of a deconvolution, Batch Normalization, and ReLU. Mathematically, we can generate the localization result $\mathbf{L} = \mathcal{B}(\mathbf{X}^a)$. While for the video-level detection result, we have $\mathbf{C} = \mathcal{C}(\mathbf{X}^a)$, where $\mathcal{C}$ is a MLP-based classifier with adaptive pooling and fully connected layers.    

Let us denote the ground-truth of localization mask and classification results as $\mathbf{L}_g$ and $\mathbf{C}_g$ respectively. The whole ST-DDL network is trained with the supervision of    

\begin{equation}
	\mathcal{L}(\mathbf{L}, \mathbf{C}, \mathbf{L}_g, \mathbf{C}_g) =  - \mathcal{L}_b(\mathbf{C}, \mathbf{C}_g) - \frac{1}{\hat{H}\hat{W}} \sum_{x = 1}^{\hat{H}} \sum_{y = 1}^{\hat{W}} \mathcal{L}_b(\mathbf{L}{(x, y)}, \mathbf{L}_g{(x, y)}),
\end{equation}
where $\mathbf{L}{(x, y)}$ (similarly for $\mathbf{L}_g{(x, y)}$) represents the $(x, y)$th entry of $\mathbf{L}$, and
\begin{equation}
    \mathcal{L}_b(\mathbf{x}, \mathbf{y}) = \mathbf{y} \log \mathbf{x} + (1-\mathbf{y}) \log (1 - \mathbf{x})
\end{equation}
is the binary cross-entropy (BCE) loss. 

Upon the presentation of the network architecture and the FA module, we are now ready to give the details of the AMM algorithm for extracting the motion clues.  

\subsection{Anchor-Mesh Motion}\label{sec:motion}
Recall that, given a series of faces $\{\CF_i\}_{i=1}^I$, the target of AMM is to obtain the motion $\CM_i$ between each pair of adjacent faces $\CF_i$ and $\CF_{i+1}$ via determining a transformation $\CT_i:\CF_i \rightarrow \CF_{i+1}$. Specifically, $\forall \bx \in \CF_i$, the transformation $\CT_i$ maps the point $\bx$ into the subsequent face $\CF_{i+1}$. The motion $\CM_i$ thus can be naturally derived as
\begin{equation}\label{eq:Mdefinition} 
\begin{aligned}
	& \CM_i = \Big\{\bx - \CT_i(\bx) \Big\}_{\bx \in \CF_i}.
\end{aligned}
\end{equation}
With (\ref{eq:Mdefinition}), the mission now reduces to determine the transformation $\CT_i$. Ideally, we can resort to solving the following optimization problem:
\begin{equation} \label{eq:main_opt}
	\mathop{\min}\limits_{\CT_i \in \mathcal{H}} \Big\|\CT_i(\CF_i) - \CF_{i+1}\Big\|_2,
\end{equation} 
where $\mathcal{H}$ is the hypothesis space of the transformation $\CT_i$.  However, the hypothesis space could be very large, making the optimization intractable. In addition, the massive number of pixels in a frame also leads to heavy computational cost, especially for nowadays popular high-definition videos.  

To alleviate these problems, we propose a constructive solution for the optimization problem (\ref{eq:main_opt}). Our solution is motivated by the property that a manifold (a face in our problem) can be approximated by the union of non-overlapping triangles, which is called the \textit{triangularization} of a manifold \cite{manifold2011}. An illustrative example is given in Fig \ref{fig:motion_cmp} (c). With such an approximation, the problem of determining the transformation between two face manifolds thus can be relaxed to finding the transformation between triangles in the two manifolds, with much reduced complexity.


Formally, we first generate $K$ triangular meshes of each frame based on pre-defined anchors. In this work, we utilize a state-of-the-art method, RetinaFace \cite{retinaface}, to calibrate these anchors. The triangularization of the face manifolds $\CF_{i}$ and $\CF_{i+1}$ are denoted by $\BF_i = \{F^{k}_i\}_{k=1}^{K}$ and $\BF_{i+1} = \{F^{k}_{i+1}\}_{k=1}^{K}$, where $F^{k}_i$ (similarly for $F^{k}_{i+1}$) is a triangle where the horizontal and vertical coordinates of its vertexes (anchors) are represented by $\ba^{k}_{i},\bb^{k}_{i}$, and $\bc^{k}_{i}$, respectively. Our goal of extracting motions now reduces to construct a series of mapping $\CA^k_i$ between every pair of triangles $F^{k}_{i}$ and $F^{k}_{i+1}$ ($k=1,...,K$). According to the \textit{collinearity} \cite{collinearity}, the transformation between two triangles should be affine or linear. By noting that every linear mapping is also an affine transformation, we only need to calculate the affine transformation $\CA^k_i$ between triangles $F^{k}_{i}$ and $F^{k}_{i+1}$. 

Generally, an affine transformation is composed of a linear mapping and a translation. For transforming triangles, the linear part can be represented by a rotation matrix derived from the cross-covariance matrix of their vertexes \cite{horn1987closed,coutsias2004using}. More precisely, the cross-covariance matrix between the vertexes of the two triangles $F^{k}_{i}$ and $F^{k}_{i+1}$ is $\mathbf{H} = \{ \bar{\ba}^{k}_i, \bar{\bb}^{k}_i, \bar{\bc}^{k}_i\}^T\{ \bar{\ba}^{k}_{i+1}, \bar{\bb}^{k}_{i+1}, \bar{\bc}^{k}_{i+1}\}$,
where 
\begin{equation}
\begin{aligned}
	 \{ \bar{\ba}^{k}_i, \bar{\bb}^{k}_i, \bar{\bc}^{k}_i\} &= \{ {\ba}^{k}_i, {\bb}^{k}_i, {\bc}^{k}_i\} - \frac{1}{3} ({\ba}^{k}_i+ {\bb}^{k}_i+ {\bc}^{k}_i), \\
	 \{ \bar{\ba}^{k}_{i+1}, \bar{\bb}^{k}_{i+1}, \bar{\bc}^{k}_{i+1}\} &= \{ {\ba}^{k}_{i+1}, {\bb}^{k}_{i+1}, {\bc}^{k}_{i+1}\} - \frac{1}{3} ({\ba}^{k}_{i+1}+ {\bb}^{k}_{i+1}+ {\bc}^{k}_{i+1})
\end{aligned}
\end{equation}
denote the centralized vertexes. Then the linear mapping of $\CA^k_i$ could be induced as
\begin{equation}
	\mathbf{R}^k_i = (\mathbf{H}^T\mathbf{H})^{\frac{1}{2}}\mathbf{H}^{-1}.
\end{equation}
Upon having the rotation matrix $\mathbf{R}^k_i$, the translation of $\CA^k_i$ can be naturally determined by shifting the centroid of the rotated triangle $F^{k}_{i}$ to that of the triangle $F^{k}_{i+1}$. Specifically, the translation matrix of $\CA^k_i$ is given by
\begin{equation}
	\mathbf{O}^{k}_i = -\frac{1}{3} \mathbf{R}^k (\bar{\ba}^{k}_{i}, \bar{\bb}^{k}_{i}, \bar{\bc}^{k}_{i})  +  \frac{1}{3} (\bar{\ba}^{k}_{i+1}, \bar{\bb}^{k}_{i+1}, \bar{\bc}^{k}_{i+1}). 
\end{equation}
Hence, the affine transformation $\CA^k_i$ between $F^{k}_i$ and $F^{k}_{i+1}$ can be expressed as 
\begin{equation}\label{eq:app_opt}
	\CA^{k}_i(\bm{x}) = \mathbf{R}^{k}_i \bm{x} + \mathbf{O}^{k}_i,
\end{equation}
for $\forall \bm{x} \in F^{k}_i$, in which $k=1,...,K$. Eventually, the motion $\CM_i$ based on the affine transformation $\CA^k_i$ can be derived 
\begin{equation}
	\CM_i = \bigcup\limits_{k=1}^{K} \Big\{\bx - \CA^{k}_i(\bx)\Big\}_{\bx \in F^k_i}.
\end{equation}

Prior to ending this subsection, we briefly discuss the theoretical justification for the approximation precision of the proposed transformation $\CA_i:\CF_i \rightarrow \CF_{i+1}$, composed of the designed affine transformations $\CA^k_i$'s in (\ref{eq:app_opt}), with respect to the optimal transformation $\CT_i$ in (\ref{eq:main_opt}). It can be shown that the distance between the transformation $\CA_i$ and the optimal $\CT_i$ is tightly bounded as follows
\begin{equation} \label{eq:upper_bound}
\|\CT_i-\CA_i \|^2_2 \leq \frac{2\lambda}{|{\CV}^c_i|^2}  + L_{i+1}A(\BF_i),
\end{equation}
where $\lambda>0$ is a constant, $|{\CV}^c_i|$ denotes the number of vertexes on the contour of the face $\CF_{i}$ (i.e., the anchors on the contour of $\CF_i$), $A(\mathbf{F}_{i})$ is the area of the mesh $\BF_i$, and $L_{i+1}$ is the length of the longest side among all triangles in the mesh $\BF_{i+1}$. The upper bound in (\ref{eq:upper_bound}) tends to zero as the increase of $|{\CV}^c_i|$ and the decrease of $L_{i+1}$. This tendency, on the one hand, indicates that more anchors on the contour of $\CF_i$ lead to more accurate transformation $\CA_i$. On the other hand, it suggests that we can enhance the approximation precision via constructing a fine-grained mesh $\mathbf{F}_{i+1}$, i.e., a large $K$, so as to avoid a triangle $F^{k}_{i+1}$ with an overlarge side length $L_{i+1}$. Guiding by these two principles, we set $|{\CV}^c_i|$ to 68, which is the maximal number of anchors supported by RetinaFace \cite{retinaface}, and empirically determine $K=90$. Under this setting, the upper bound in (\ref{eq:upper_bound}) becomes 0.24, which is very tight. The proof and more justifications can be found in the appendix.

\section{\texttt{ManualFake} Dataset}
\begin{figure}[ht]
	\centering
	\subfigure{
		\includegraphics[width = 0.8\textwidth]{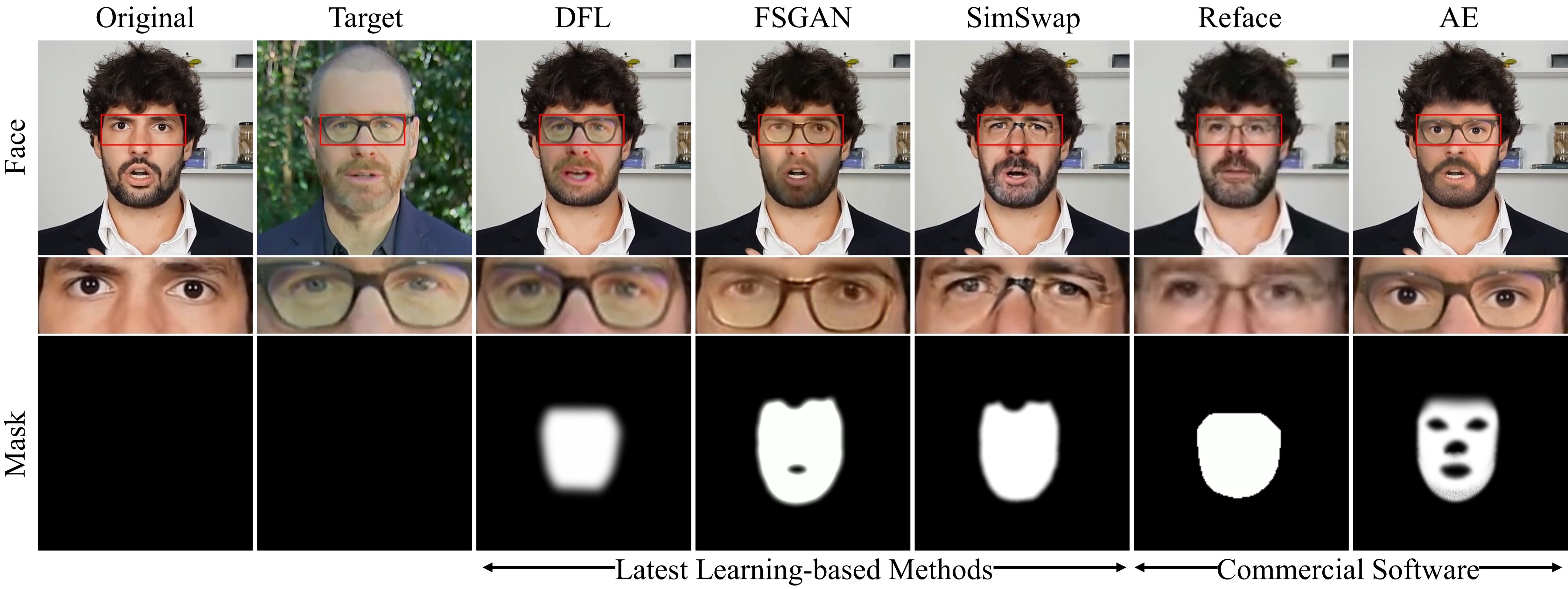}}
	\caption{Preview of \texttt{ManualFake}. It includes the forgeries produced by the latest deep methods (DFL, FSGAN, SimSwap), and commercial software (Reface and AE). Best viewed with zoom-in.}
	\label{fig:dataset}
	\vspace{-2mm}
\end{figure}
We construct a new Deepfake dataset, \texttt{ManualFake}, that involves forgeries produced by not only latest learning-based methods (DFL \cite{dfl}, FSGAN \cite{fsgan}, and SimSwap \cite{simswap}), but also by two widely-adopted commercial software (After Effects (AE) \cite{AdobeAE} from Adobe and Reface \cite{reface} from NeoCortext). In fact, AE and Reface have been used by many video editing professionals to generate high-quality fake videos. Also, noting that OSNs are the dominating channels for spreading out Deepfake videos, we also include the transmitted versions through four most popular OSN platforms: Facebook, Whatsapp, Tiktok, and Wechat. These OSN-transmitted versions are created by manually uploading the original manipulated videos and downloading their processed ones. Eventually, we collect 6000 videos (1000 pristine, 1000 untransmitted fake, and 4000 OSN-transmitted fake) with resolutions ranging from 360p up to 1080p, together with pixel-level segmentation masks indicating the forged regions. An example of our dataset \texttt{ManualFake} is given in Fig. \ref{fig:dataset}. As can be observed, when synthesizing glasses and beard on a face, using commercial software AE could lead to much better visual quality than learning-based methods. This explains why we need to involve commercial software produced forgeries, in addition to the ones generated by learning-based methods.

\begin{table}[t]
	\caption{Video quality evaluations on different Deepfake datasets. The highest value among these datasets is \textbf{bold} while the second-best is \underline{underlined}.}
	\centering
	\label{tab:ourset_analysis}
	\scalebox{0.75}{
		\begin{tabular}{l|c|c|c|c||ccccc}
			\hline
			\hline
			\multirow{2}{*}{Metric} & \multirow{2}{*}{\texttt{FF++} \cite{ff++}} & \multirow{2}{*}{\texttt{CelebDF} \cite{CelebDF}} & \multirow{2}{*}{\texttt{FFIW} \cite{ffiw}} & \texttt{ManualFake} & \multicolumn{5}{c}{Sub-classes of \texttt{ManualFake}}\\
			& & & & Mean & DFL & FSGAN & SimSwap & Reface & AE \\
			\hline
			EQFace \cite{EQFace} & \underline{.6725} & .6318 & .5473 & \textbf{.7355} & .7254 & .6979 & .8120 & .7237 & .7185 \\
			FaceQnet \cite{FaceQnet} & \underline{.3682} & .3437 & .3186 & \textbf{.4125} & .4108 & .4314 & .3838 & .4103 & .4262\\
			\hline
			\hline
		\end{tabular}
	}
\end{table}

To quantitatively measure the generation quality of the Deepfake videos in \texttt{ManualFake}, we adopt two non-reference face recognition metrics, namely EQFace \cite{EQFace} and FaceQnet \cite{FaceQnet} (higher the better), according to \cite{GANFaceQnet}. Table \ref{tab:ourset_analysis} shows the comparisons on datasets \texttt{FF++} \cite{ff++}, \texttt{CelebDF} \cite{CelebDF}, \texttt{FFIW} \cite{ffiw} and our \texttt{ManualFake}. To provide fine-grained comparison within \texttt{ManualFake}, we also give the score of each sub-class, together with the mean score. As can be seen, \texttt{ManualFake} achieves an average of 0.7355 EQFace and 0.4125 FaceQnet, which are the highest scores among all the competing datasets. More precisely, our dataset outperforms the second best one with 6.3\% and 4.4\% improvements, in terms of EQFace score and FaceQnet score, respectively. We should also emphasize that all sub-classes in \texttt{ManualFake} surpass the existing datasets \cite{ff++, CelebDF, ffiw}. For more featured details about \texttt{ManualFake}, please refer to the supplementary material.

\section{Experiments}\label{sec:exp}
\textbf{Datasets.} Similar to \cite{lip2021, ftcn2021}, we train our model on the dataset \texttt{FF++} \cite{ff++} (high-quality version), which contains 1000 pristine videos and 4000 fake videos manipulated by Deepfake \cite{deepfake}, FaceSwap \cite{faceswap}, Face2Face \cite{face2face}, and NeuralTexture \cite{neuraltexture}. To concurrently evaluate the detection and localization of different Deepfake forensic algorithms, we adopt the widely-used datasets \texttt{DFD} \cite{dfd} and \texttt{FFIW} \cite{ffiw}. We also involve our newly proposed dataset \texttt{ManualFake} to expand the data diversity. Note that there is \textbf{NO} overlap between the training and testing datasets, better simulating the real situation and evaluating the generalization performance of these Deepfake forensic algorithms.

\textbf{Comparative Methods and Metrics.} To better evaluate the detection and localization performance, we compare our model with the following state-of-the-art detection methods, FlowCNN \cite{flowcnn}, LipNet \cite{lip2021}, and FTCN \cite{ftcn2021}, as well as the localization approaches, WaveNet \cite{wavenet}, FakeLocator \cite{fakelocator} and CFFs \cite{cffs}. Three commonly-used metrics are adopted, namely, the AUC, the F1, and the IoU. Following the settings in \cite{ff++, lip2021}, all the evaluations are based on the same number of frames. Please refer to the appendix for more details on the calculations of these metrics.

\textbf{Implementation Details.} Our proposed ST-DDL network is implemented using the PyTorch framework and the experiments are performed on an NVIDIA Tesla A100 GPU. We supervise the training by AdamW \cite{adamw} optimizer and Cosine Annealing \cite{adamw} learning strategy. The minibatch size is set to 8, and input images are cropped and resized to $512 \times 512$. As for the FA, we set the self-attention heads and the hidden dimension of linear projection to 16 and 512, respectively. In the training process, we randomly introduce various data augmentations, such as flipping and compression, so as to improve the robustness of the model \cite{easyspot2020}. To embrace the concept of reproducible research, the code of our paper and the collected dataset are made available at \textbf{https://github.com/HighwayWu/ST-DDL}, serving as a useful resource to our research community for fighting against the Deepfake.

\subsection{Quantitative Results}
The quantitative comparisons in terms of the video-level F1 and AUC, pixel-level F1 and IoU (higher the better) are presented in Table \ref{tab:quant}. Under these evaluation criteria, our proposed method consistently outperforms the corresponding second-place competitor on all three testing datasets. Numerically, we surpass the FTCN \cite{ftcn2021} by 4.1\% and 3.7\% video-F1 on datasets \texttt{DFD} \cite{dfd} and \texttt{ManualFake}, respectively, and even exceed the CFFs \cite{cffs} by 8.9\% on \texttt{FFIW} \cite{ffiw}. Regarding the localization performance, our ST-DDL achieves 3.3\%, 3.2\%, and 4.1\% pixel-IoU gains compared to FakeLocator \cite{fakelocator} on the three testing datasets, respectively. Here, for \texttt{ManualFake}, we only adopt the pristine videos and the untransmitted ones, while deferring the evaluations of OSN-transmitted \texttt{ManualFake} in Sec.~\ref{sec:robustness}. Compared to video-level metrics, all methods achieve inferior performance under pixel-level metrics, indicating that forgery localization is a challenging task and worthy of our long-term research efforts.

\begin{table}[t]
	\caption{Quantitative comparisons by using video-level F1 and AUC ($\mathrm{V}_\mathrm{F1}$ and $\mathrm{V}_\mathrm{AUC}$), pixel-level F1 and IoU ($\mathrm{P}_\mathrm{F1}$ and $\mathrm{P}_\mathrm{IoU}$) as criteria. For each column, the highest value is \textbf{bold}, while the second-best is \underline{underlined}, and ``-'' indicates not applicable.}
	\centering
	\label{tab:quant}
	\scalebox{0.8}{
		\begin{tabular}{l|cccc|cccc|cccc}
			\hline
			\hline
			\multirow{2}{*}{Method} & \multicolumn{4}{c}{\texttt{DFD} \cite{dfd}} & \multicolumn{4}{c}{\texttt{FFIW} \cite{ffiw}} & \multicolumn{4}{c}{\texttt{ManualFake}} \\
			\cline{2-13}
			& $\mathrm{V}_\mathrm{F1}$ & $\mathrm{V}_\mathrm{AUC}$ & $\mathrm{P}_\mathrm{F1}$ & $\mathrm{P}_\mathrm{IoU}$ & $\mathrm{V}_\mathrm{F1}$ & $\mathrm{V}_\mathrm{AUC}$ & $\mathrm{P}_\mathrm{F1}$ & $\mathrm{P}_\mathrm{IoU}$ & $\mathrm{V}_\mathrm{F1}$ & $\mathrm{V}_\mathrm{AUC}$ & $\mathrm{P}_\mathrm{F1}$ & $\mathrm{P}_\mathrm{IoU}$ \\
			\hline
			FlowCNN \cite{flowcnn} & .482 & .596 & - & - & .333 & .502 & - & - & .317 & .514 & - & - \\
			LipNet \cite{lip2021} & .607 & .788 & - & - & .594 & .625 & - & - & .675 & .685 & - & - \\
			FTCN \cite{ftcn2021} & \underline{.692} & .870 & - & - & .596 & .700 & - & - & \underline{.691} & .732 & - & - \\
			WaveNet \cite{wavenet} & .556 & .844 & .623 & .520 & .580 & .743 & .354 & .285 & .653 & .727 & .506 & .412 \\
			FakeLocator \cite{fakelocator} & .616 & \underline{.876} & \underline{.718} & \underline{.601} & .615 & .735 & \underline{.618} & \underline{.521} & .660 & \underline{.738} & \underline{.519} & \underline{.431} \\
			CFFs \cite{cffs} & .576 & .836 & .681 & .575 & \underline{.654} & \underline{.798} & .461 & .388 & .614 & .715 & .486 & .403 \\
			ST-DDL (Ours) & \textbf{.733} & \textbf{.912} & \textbf{.741} & \textbf{.634} & \textbf{.743} & \textbf{.855} & \textbf{.656} & \textbf{.554} & \textbf{.728} & \textbf{.780} & \textbf{.565} & \textbf{.472} \\
			\hline
			\hline
		\end{tabular}
	}
\end{table}

\begin{figure}[t]
	\centering
	\subfigure{
		\includegraphics[width = 1\textwidth]{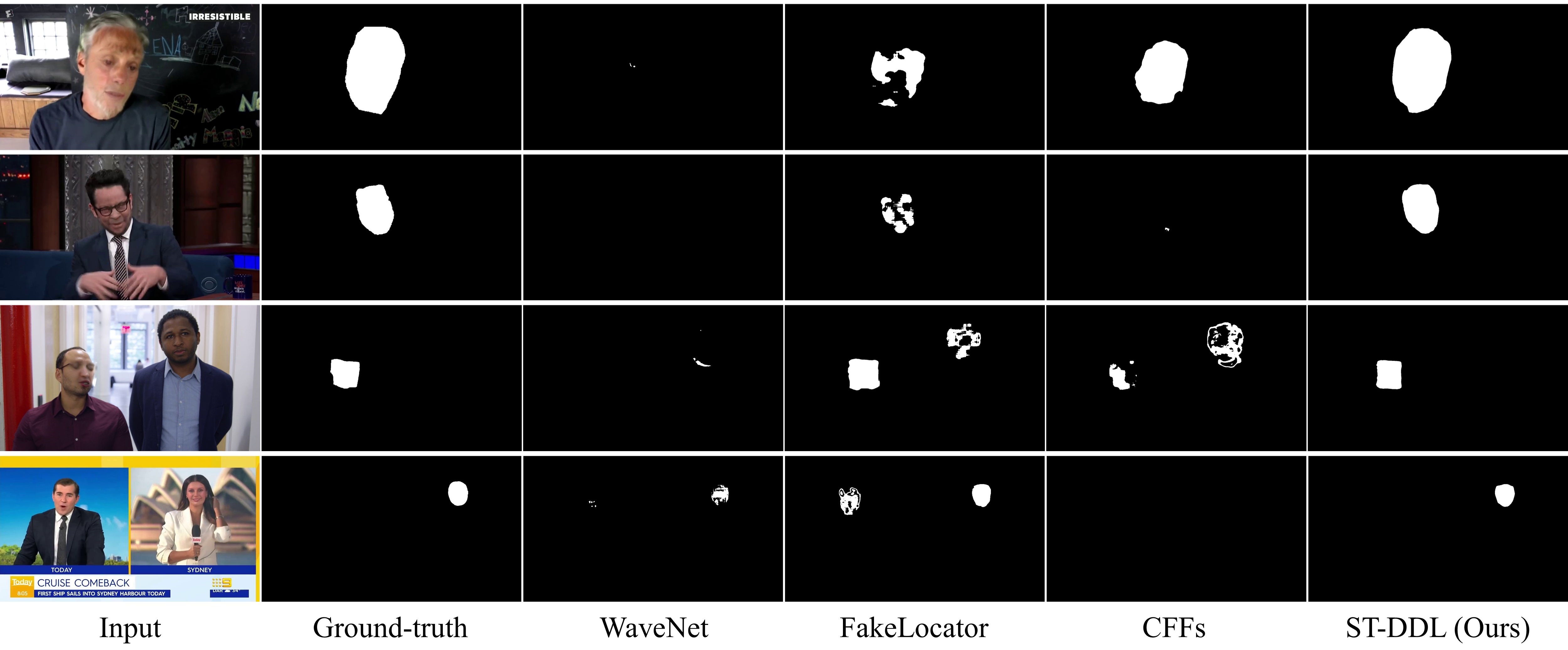}}
	\caption{Comparisons of Deepfake localization. The images from left to right are input, ground-truth, localization results of WaveNet \cite{wavenet}, FakeLocator \cite{fakelocator}, CFFs \cite{cffs} and our ST-DDL, respectively.}
	\label{fig:vis_cmp}
	\vspace{-2mm}
\end{figure}

\begin{figure}[t]
	\centering
	\subfigure{
		\includegraphics[width = 0.8\textwidth]{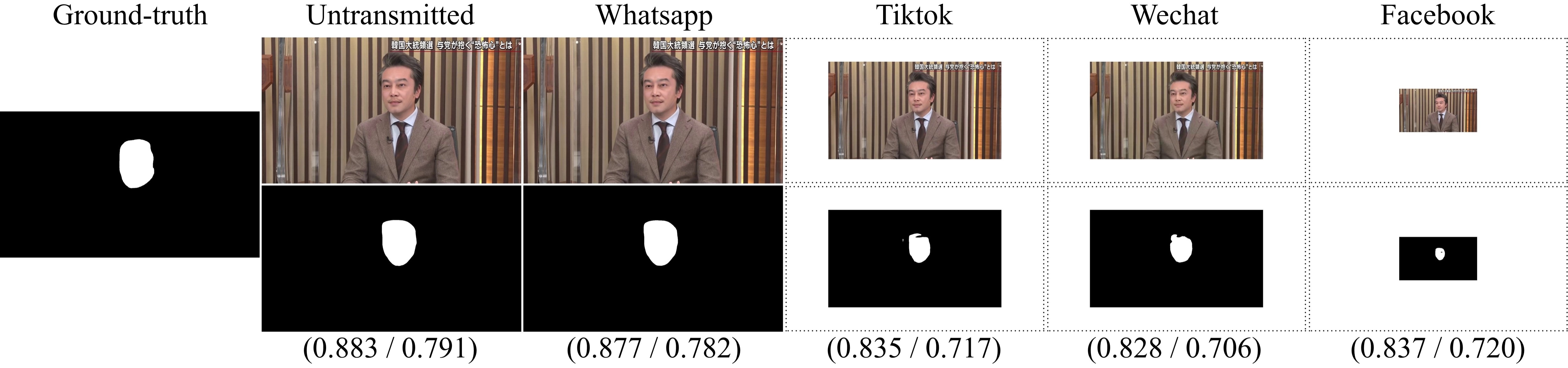}}
	\renewcommand{\thefigure}{7}
	\caption{Localization results of ST-DDL under the OSN transmissions, where the pixel-level F1/IoU scores are provided. Note that many OSN transmissions result in reduction of the video resolution.}
	\label{fig:osn_loc}
\end{figure}



\subsection{Qualitative Comparisons}
In addition to the quantitative comparisons, we also compare different methods qualitatively. In Fig. \ref{fig:vis_cmp}, we show the localization results on several representative examples from testing dataset \texttt{DFD} \cite{dfd}, \texttt{FFIW} \cite{ffiw}, and our \texttt{ManualFake}. It can be observed that the competitors sometimes have mislocalization in different regions and cannot maintain consistent performance. For instance, FakeLocator \cite{fakelocator} has difficulties in accurately locating the forged area, such as the cases in the first and second rows, while WaveNet \cite{wavenet} and CFFs \cite{cffs} sometimes tend to neglect frontal or minor faces, leading to missed judgments in rows 2-4. In contrast, our method can learn more distinctive spatial and temporal features, and thereby generate more precise localization results. Due to the space limit, more qualitative results are given in the supplementary file.

\subsection{Robustness Evaluation}\label{sec:robustness}
We would also like to evaluate the robustness of different Deepfake forensic methods under the transmission over various OSNs. This evaluation is very essential in practical scenarios, since OSNs are the dominating channels for disseminating Deepfake videos, and the inevitable lossy operations in OSNs could have huge impact on the detection and localization performance \cite{wei2016osn,wei2020osn,wei_optimal2021, dresden,wu2022robust}. Fig.~\ref{fig:osn_cmp} presents the performance of different models not only on the 1000 untransmitted fake videos, but additionally on the 4000 OSN-transmitted versions including the most accessible platforms Facebook, Whatsapp, Tictok, and Wechat. The results show that all methods suffer from performance degradation, due to the lossy operations applied on OSNs, particularly for Facebook data. The reason may be that Facebook conducts substantial quality compression and video frame insertion, e.g., the maximum bit rate compression is over 95\%, and frame per second (FPS) is fixed to 30. Interestingly, we find that Whatsapp is the most forensics-friendly platform, with much larger allowable bitrates (e.g., >1087 kbit/s) and no additional lossy operations. An illustrative example of localizing OSN-transmitted forgery is presented in Fig.~\ref{fig:osn_loc}. More examples are given in the appendix.

\begin{table}[t]
    \hspace{-2mm}
	\begin{minipage}{0.6\linewidth}
    	\vspace{7mm}
		\centering
		\includegraphics[width = 0.9\linewidth]{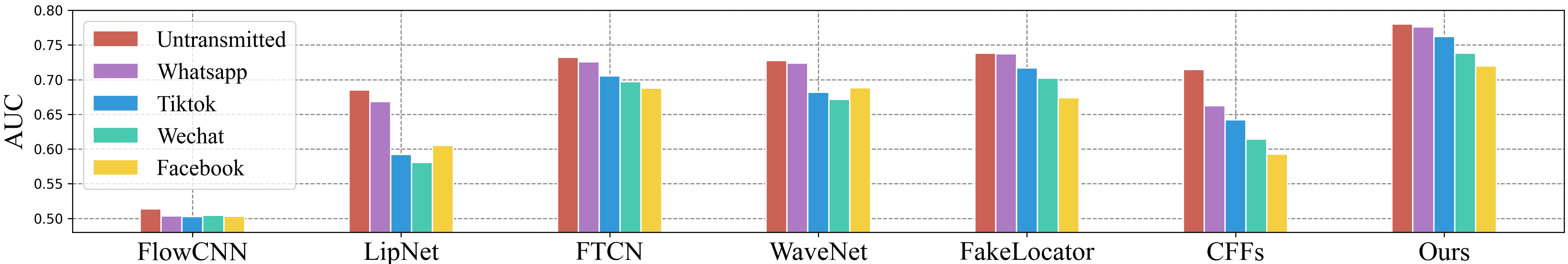}
		\renewcommand{\tablename}{Figure}
		\renewcommand{\thetable}{6}
		\caption{Robustness against OSN transmissions.}
		\label{fig:osn_cmp}
	\end{minipage}
	\begin{minipage}{0.4\linewidth}
		\caption{Ablation studies regarding different architectures on \texttt{Face2Face}.}
		\label{tab:ablation}
		\centering
		\scalebox{0.5}{
			\begin{tabular}{l|cccc|c}
				\hline
				\hline
				Architecture & $\mathrm{P}_\mathrm{F1}$ & $\mathrm{P}_\mathrm{IoU}$ & $\mathrm{V}_\mathrm{F1}$ & $\mathrm{V}_\mathrm{AUC}$ & Mean \\
				\hline
				\#1 RGB & .682 & .605 & .781 & .915 & .746 \\
				\#2 RAFT \cite{raft} & .649 & .573 & .596 & .682 & .625 \\
				\#3 AMM & .678 & .605 & .651 & .750 & .671 \\
				\#4 RGB + AMM (Conv.) & .724 & .667 & .875 & .968 & .808 \\
				\#5 RGB + AMM (FA) & .771 & .712 & .897 & .977 & .839 \\
				\hline
				\hline
			\end{tabular}
		}
	\end{minipage}\hfill
	\vspace{-2mm}
\end{table}

\subsection{Ablation Studies}
For analyzing how the designed AMM features and FA module contribute to the model performance, we construct five variants and present their performance in Table \ref{tab:ablation}. Specifically, variant \#1 contains the RGB encoder alone, while \#2 and \#3 utilize only motion encoder, in which RAFT motion \cite{raft} and ours are used as inputs, respectively. Variants \#4 and \#5 are to compare the effectiveness of the convolutional concatenation and the FA module. From the results, \#1 architecture performs satisfactorily in general (91.5\% AUC) by using only spatial RGB features. Compared with RAFT motion \cite{raft} that can hardly learn useful information from, our AMM can provide much more temporal forensic clues, resulting in a 6.8\% AUC gain. However, the relatively inferior performance (68.2\% and 75.0\% AUC scores) suggests that motions alone are not suitable for completing forensic tasks. By concatenating the RGB and AMM features, \#4 variant achieves a 6.2\% improvement over \#1 on the four metrics in Table \ref{tab:ablation}. Finally, through utilizing a delicate FA module, variant \#5 could much better exploit the spatial and temporal forensic evidences, leading to a further 3.1\% average gain.

\subsection{Limitations and Ethics Discussion}\label{sec:limit}
Although our method performs well in the above experiments, it still has room for further improvements. Similar to the methods \cite{lip2021} exploiting temporal features, our ST-DDL may suffer from a performance decline in detecting videos with interpolated frames, in which case the AMM algorithm can only extract rather limited motion clues. Such a limitation could be alleviated by introducing more video sequences during training. Additionally, utilizing data augmentation methods, e.g., adversarial training, may further improve the robustness of the model to combat severe OSN interference.       

We also address some potential ethical concerns of our work: 1) The proposed \texttt{ManualFake} is primarily intended to advance forensic research, so as to alleviate the malicious intent of Deepfake. 2) \texttt{ManualFake} is produced by non-public methods and thereby not reproducible. 3) The video contents in \texttt{ManualFake} are legal and ethical (e.g., from official news). 4) We develop a new forensic algorithm to fight against the Deepfake.

\section{Conclusion}
We propose a novel ST-DDL network for simultaneously detecting the originality and locating the forged regions of Deepfake videos. Specifically, an AMM algorithm is delicately designed to extract distinctive facial motion features for Deepfake forensics. Furthermore, a FA module is equipped for the interaction of spatial and temporal features, significantly boosting the forgery detection and localization performance. Extensive experimental results are provided to validate the superiority of our ST-DDL. In addition, to promote the future development of Deepfake forensics, we build a new public forgery dataset, having many unique features.     

{
	\small
	\bibliographystyle{IEEEtran}
	\bibliography{ref}
}

\end{document}